\documentclass[sigconf]{acmart}
\usepackage{subfigure}
\usepackage{hyperref}
\AtBeginDocument{%
  \providecommand\BibTeX{{%
    \normalfont B\kern-0.5em{\scshape i\kern-0.25em b}\kern-0.8em\TeX}}}

\copyrightyear{2022}
\acmYear{2022}
\setcopyright{acmlicensed}\acmConference[SIGIR '22]{Proceedings of the 45th International ACM SIGIR Conference on Research and Development in Information Retrieval}{July 11--15, 2022}{Madrid, Spain}
\acmBooktitle{Proceedings of the 45th International ACM SIGIR Conference on Research and Development in Information Retrieval (SIGIR '22), July 11--15, 2022, Madrid, Spain}
\acmPrice{15.00}
\acmDOI{10.1145/3477495.3531848}
\acmISBN{978-1-4503-8732-3/22/07}

\begin{document}

\fancyhead{}
\title{GraphAD: A Graph Neural Network for Entity-Wise Multivariate Time-Series Anomaly Detection}

\author{Xu Chen}
\email{sylover@pku.edu.cn}
\affiliation{%
  \institution{School of Artificial Intelligence, Peking University}
  \city{Beijing}
  \country{China}
}

\author{Qiu Qiu}
\email{ricklovelisa@126.com}
\affiliation{%
  \institution{Alibaba DAMO Academy}
  \city{Hangzhou}
  \country{China}
}

\author{Changshan Li}
\email{lics16@mails.tsinghua.edu.cn}
\affiliation{%
  \institution{Alibaba DAMO Academy}
  \city{Hangzhou}
  \country{China}
}

\author{Kunqing Xie}
\email{kunqing@pku.edu.cn}
\affiliation{%
  \institution{School of Artificial Intelligence, Peking University}
  \city{Beijing}
  \country{China}
}

\renewcommand{\shortauthors}{Xu Chen and Qiu Qiu, et al.}

\begin{abstract}
    In recent years, the emergence and development of third-party platforms have greatly facilitated the growth of the Online to Offline (O2O) business. However, the large amount of transaction data raises new challenges for retailers, especially anomaly detection in operating conditions. Thus, platforms begin to develop intelligent business assistants with embedded anomaly detection methods to reduce the management burden on retailers. Traditional time-series anomaly detection methods capture underlying patterns from the perspectives of time and attributes, ignoring the difference between retailers in this scenario. Besides, similar transaction patterns extracted by the platforms can also provide guidance to individual retailers and enrich their available information without privacy issues. In this paper, we pose an entity-wise multivariate time-series anomaly detection problem that considers the time-series of each unique entity. To address this challenge, we propose GraphAD, a novel multivariate time-series anomaly detection model based on the graph neural network. GraphAD decomposes the Key Performance Indicator (KPI) into stable and volatility components and extracts their patterns in terms of attributes, entities and temporal perspectives via graph neural networks. We also construct a real-world entity-wise multivariate time-series dataset from the business data of Ele.me. The experimental results on this dataset show that GraphAD significantly outperforms existing anomaly detection methods.
\end{abstract}



\maketitle

\section{introduction}
With the development of e-commerce, the scale of the Online to Offline (O2O) business has been remarkably promoted, raising a great demand for real-time and secure transactions. Numerous third-party trading platforms have emerged to facilitate O2O business. Ele.me, for example, is one of the largest catering and food delivery service platforms in China. So far, it has covered 2,000 cities with 1.3 million retailers joining and 260 million users\footnote{The data is obtained from the official website \url{https://www.ele.me/}.}. With the online trading capacity of such platforms, consumers are able to purchase a variety of products, and retailers can sell goods to consumers living in remote areas conveniently. However, the intricate behaviors of customers such as order cancellation, complaints and refunds will generate a considerable amount of extra transaction data, thus severely affecting the self-assessment of retailers on their operating conditions. Therefore, these platforms have started to develop intelligent business assistants that provide information and business suggestions for retailers. Anomaly detection is one of the most critical functions 
because it provides signals for abnormal operating conditions so that retailers can avoid potential financial losses.

The existing anomaly detection methods embedded in intelligent business assistants are mainly designed for univariate/multivariate time-series. Univariate time-series anomaly detection methods like ARIMA\cite{box2015time}, STL\cite{robert1990stl}, RobustSTL\cite{wen2019robuststl} and wavelet analysis\cite{wen2021robustperiod} directly analyze the pattern of Key Performance Indicator (KPI), which cannot handle the multivariate data and then reduce the available information. Methods for multivariate time-series extract underlying correlations between KPI and other attributes. Previous studies introduce various technologies such as tree-based method Isolation Forest\cite {liu2008isolation}, empirical copula-based method COPOD\cite{li2020copod} and deep learning methods VAE\cite{an2015variational} and LSTM-AE \cite{malhotra2016lstm}. Recently, some works have begun to integrate Graph Neural Networks (GNNs) into anomaly detection \cite{deng2021graph,zhao2020multivariate}. Although these researches have achieved impressive performance in time-series anomaly detection, only temporal factors and the impact of different attributes are integrated, leaving the distinction of different entities not yet studied.

\begin{figure*}[!ht]
\centering
\subfigure[Retailers A]{
    \includegraphics[width=0.9\columnwidth]{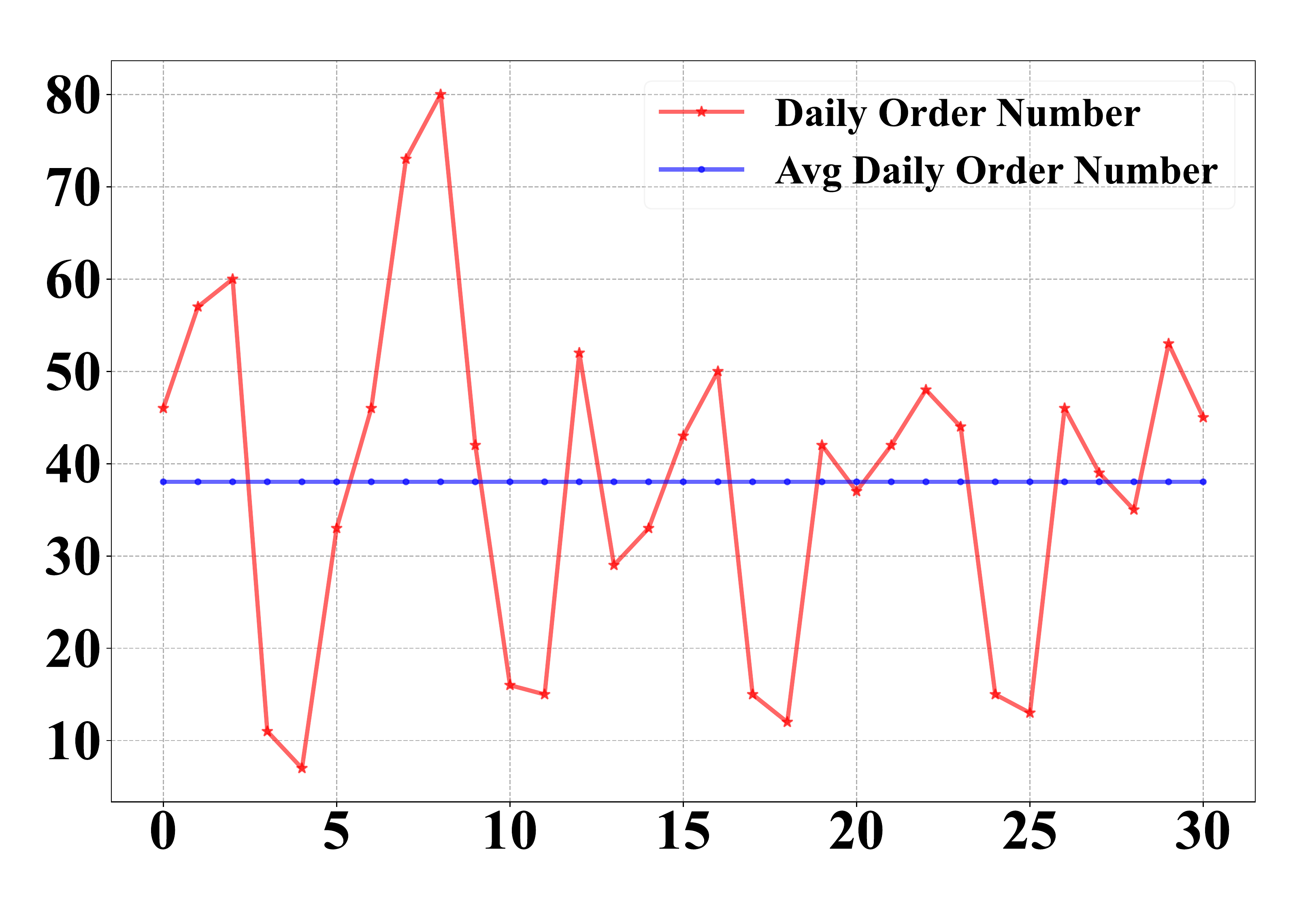}
    \label{fig:a}
}
\subfigure[Retailers B]{
    \includegraphics[width=0.9\columnwidth]{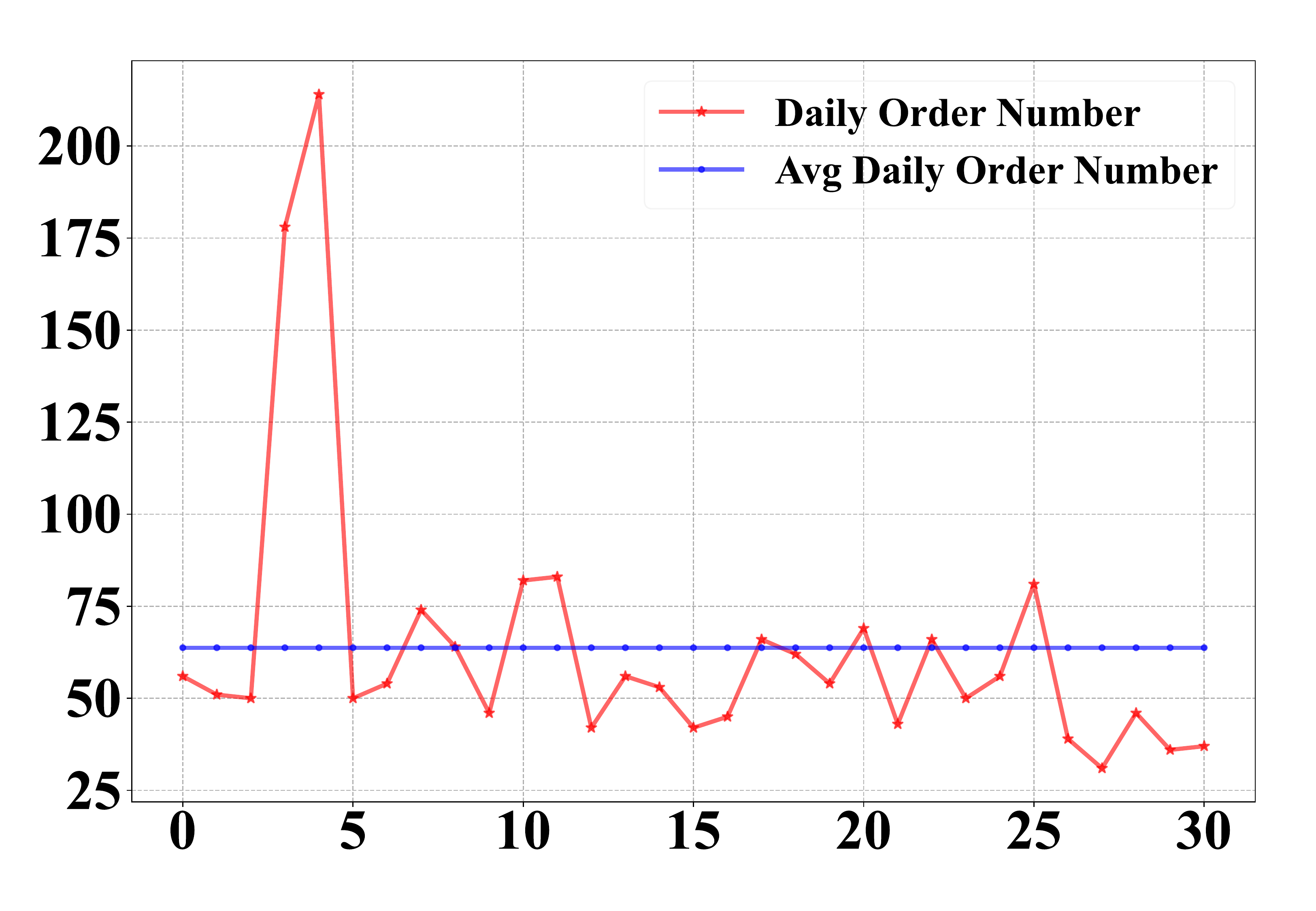}
    \label{fig:b}
}

\subfigure[Retailers C]{
    \includegraphics[width=0.9\columnwidth]{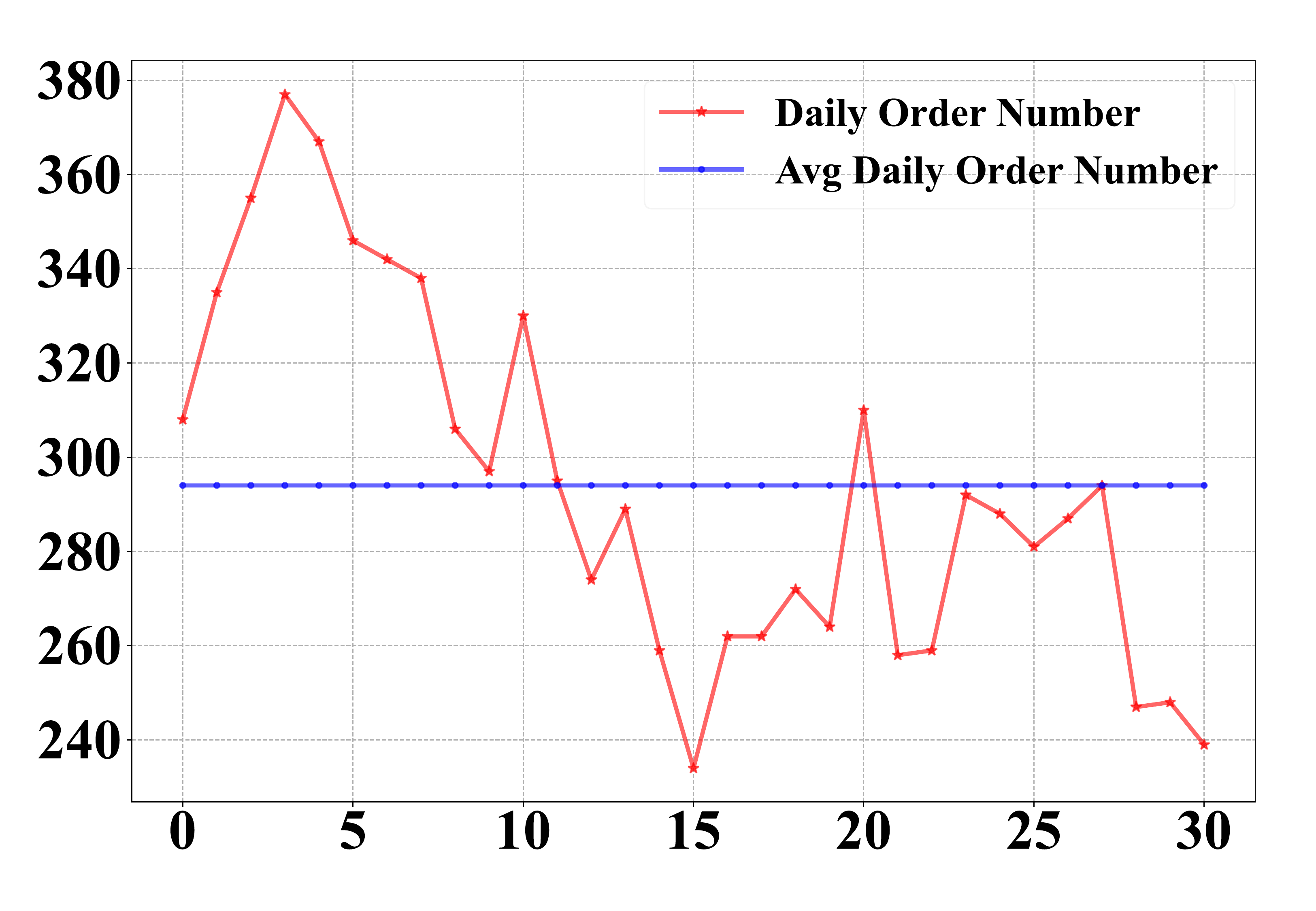}    
    \label{fig:c}
}
\subfigure[Retailers D]{
\includegraphics[width=0.9\columnwidth]{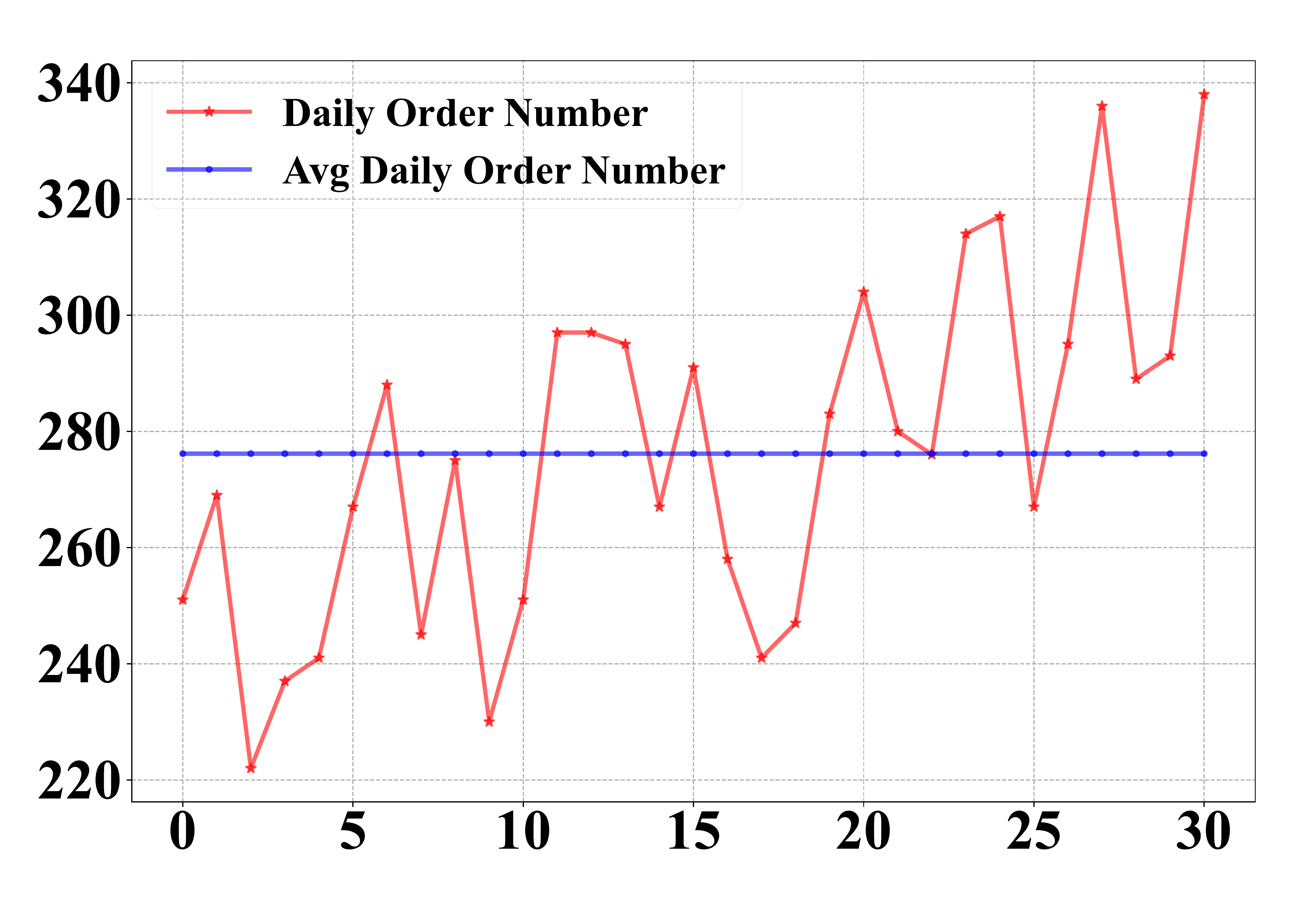}
    \label{fig:d}
}
\caption{Difference between KPI patterns across entities on Ele.me in 31 days. Entities mean retailers in this scene, and KPI corresponds to the daily order number. The vertical axis denotes the daily order number of each retailer, and the horizontal axis indicates different days.}
\label{fig:res}
\end{figure*}

In this paper, we pose a new anomaly detection problem in the O2O platform scenario, namely entity-wise multivariate time-series anomaly detection (EMTAD). The entity represents the object of interest, which refers to the retailer of Ele.me. Unlike previous studies, numerous entities in EMTAD should not be regarded as homogeneous. The primary reason is that the scale and business patterns vary significantly across entities. We randomly select four retailers (entities) and depict their 31 days daily order numbers in the red line and average daily order numbers within 31 days in the blue line. As illustrated in Fig.\ref{fig:res}, there is a significant difference in their turnover: the average daily order number of retailer A and B is less than 100, while it is around 300 for retailer C and D. The evolution patterns of daily order numbers are also diverse. The pattern stays steady for retailer A and it shows an upward trend of retailer D. In contrast, it presents distinct volatility for retailer C and even has two abnormal points for retailer B. Previous methods capture the temporal patterns or attribute correlations without distinction of entities. In other words, all retailers are viewed as 
identical and equally modeled, which neglects their unique characteristics. These methods will also face great challenges in model training due to the data variance from different entities. As a result, they are expected to recognize the common patterns only and have trouble portraying each retailer in detail. Such challenges reveal the importance and necessity of anomaly detection in EMTAD.

In the EMTAD scenario, the most vital problem is \textit{how to capture the intricate correlations between different entities?} Moreover, the inter-retailer correlations are heterogeneous and can be divided into two types. Firstly, retailers with similar static information are highly correlated. For example, compared with retailers selling desserts, patterns of retailers offering spicy food will be more similar. Secondly, the similarity of time-varying patterns reveals dynamic correlations between retailers. For instance, the transaction trends of newly started shops evolve like other incipient shops and are distinguished from shops that have opened for a long time. Thus, the choice of an appropriate method becomes the most critical step in EMTAD.

Recent years have witnessed the remarkable progress of GNN \cite{kipf2016semi,hamilton2017inductive,velivckovic2017graph,chen2020tssrgcn,chentrafficstream}, and the capacity of graphs to represent relations between objects is an essential reason for its success. Therefore, an ideal solution is to model the aforementioned correlations via graphs. In this paper, we propose a graph neural network-based model, namely GraphAD, for the EMTAD problem. GraphAD consists of three modules: a KPI decomposition (K-Decom) module, an attribute graph attention network and an entity-time graph attention network. Considering the variance of the stable component and the volatility of KPI, we design the K-Decom module to decompose KPI into the stable series and the volatile series. Attribute graph attention networks are adopted to capture the relationships between attributes of the outputs of K-Decom module, respectively. The entity-temporal graph attention network is designed for mining the correlations between different retailers from both the static and dynamic aspects. 

The contribution of this paper can be summarized as follows:
\begin{itemize}
    \item We pose an entity-wise multivariate time-series anomaly detection problem in the O2O scene. To the best of our knowledge, it is the first study attempting to distinguish different entities in this area.
    \item We elaborate on a graph neural network-based anomaly detection model, GraphAD. It investigates the difference between stable component and volatility of KPI. Both the inter-attribute and inter-retailer correlations are integrated through graph attention networks.
    \item We construct a real-world EMTAD dataset from Ele.me data. Extensive experiments show the effectiveness of GraphAD over widely applied anomaly detection methods.
\end{itemize}

\section{related work}
Time-series anomaly detection detects abnormal status given specific attribute(s) organized in time series. Previous studies can be divided into univariate anomaly detection and multivariate anomaly detection. Methods under the former setting directly analyze the pattern of the most concerned attribute, e.g., KPI, such as ARIMA\cite{box2015time}, STL\cite{robert1990stl}, RobustSTL\cite{wen2019robuststl} and wavelet analysis\cite{wen2021robustperiod}. On the latter scene, multiple attributes are simultaneously considered to detect whether KPI is abnormal or not. Isolation Forest\cite {liu2008isolation} is a tree-based machine learning method that can be applied in the unsupervised scene, while COPOD\cite{li2020copod} detects anomalies with the empirical copula. Deep learning methods like VAE\cite{an2015variational} and LSTM-AE\cite{malhotra2016lstm} further promote the performance on anomaly detection. Reconstruction probability is designed as the criterion for distinguishing outliers in VAE. Compared with AutoEncoder, LSTM-AE replaces the neural networks of the encoder and the decoder with LSTM to enhance the capacity for handling time series. 

Recent studies have begun to introduce Graph Neural Networks into anomaly detection. GDN \cite{deng2021graph} uses structure learning to model the correlations among sensors. MTAD-GAT \cite{zhao2020multivariate} adopts graph attention networks to capture dependency among time-series and attributes. However, these methods are designed for multivariate time-series, which cannot handle the difference between retailers in EMTAD.

\section{GraphAD}
In this section, we will elaborate on the details of GraphAD. Raw data will firstly be fed into KPI Decomposition Module to generate the stable and the volatility component. They will then be passed into two attribute graph attention networks (A-GAT) respectively to capture each inherent correlation across attributes. Outputs of two A-GAT will become the inputs of two entity-temporal graph attention networks (ET-GAT). The representations learned by ET-GAT will finally be concatenated for an MLP layer. In summary, two different components are decomposed from input series, and are then fed to corresponding graph attention networks. Final predictions are generated from their concatenated representations.
\subsection{KPI Decomposition Module}
Some previous anomaly detection methods usually decompose KPI and remove the periodic patterns first\cite{robert1990stl,wen2019robuststl}. The key idea is that the stable component and the volatility of KPI should be distinguished for their different evolution process. Considering the message passing procedure of GNN, the two different components will be fused by the same non-linear transformation, which smooths the volatility and causes information loss. Regarding this problem, it is natural and reasonable to decompose the KPI before propagation through GNNs. We extract the stable component $X^S$ and volatility $X^V$ from the original KPI $X$ through K-Decom module with two independent stacked fully-connected layers $f_S$ and $f_V$:
\begin{equation}
\label{eq:decompose}
\begin{split}
    X^S & = f_S\left(X,\Theta_S\right)=\sigma(X\Theta_S),\\ 	X^V & = f_V\left(X,\Theta_V\right)=\sigma(X\Theta_V),
\end{split}
\end{equation}
where $\Theta_S\in\mathbb{R}^{D\times D_S}$ and $\Theta_V\in\mathbb{R}^{D\times D_V}$ are learnable parameters of the stacked fully-connected layers. $D$ is the dimension of input series. $D_S=D_V$ is the dimension of output and $\sigma(\cdot)$ is the activation.

Inspired by information bottleneck theory\cite{shwartz2017opening,mao2021neuron}, we design two mutual information-based objectives to impose constraints on two components and obtain meaningful results. The stable one is more regular and the correlation with the KPI series should be stronger. 
Hence, we maximize their mutual information by:
\begin{equation}
\label{eq:stable}
	\max_{\mathbf{\Theta}_S} \quad I(X,X^S).
\end{equation}
On the contrary, the volatility one $X^V$ is more random and unstable, which is similar to ``noise'' from the data perspective. It suggests the necessity to reduce the mutual information between $X^V$ and $X$ as:
\begin{equation}
\label{eq:vola}
	\min_{\mathbf{\Theta}_V} \quad I(X,X^V).
\end{equation}
The computation of two mutual information is discussed in Sec. \ref{mutual_compute}.
Eq. \eqref{eq:stable} and \eqref{eq:vola} can be interpreted from another aspect that we can adopt the IB theory to ``reserve'' the informative stable component $X^S$ and ``filter out'' volatility one $X^V$. Note that two components are both valuable and are used for downstream modeling.

\subsection{Attribute Graph Attention Network}
The relations between the input attributes are rather intricate in EMTAD scenario, and external factors such as shopping carnivals of platforms can further influence their correlations. Therefore, it is necessary to dynamically capture such correlations that are too challenging to be organized as structured data. The recent development of the graph has verified its effectiveness in portraying the unstructured relations between objects. Based on its impressive capability, we model the attributes of entity $n$ at timestep $t$ as an attribute graph $G_A^{nt}$. In more detail, each node on the attribute graph represents an input attribute. For a node $v_i$, we calculate time-series similarity between $v_i$ and the rest nodes $v_j \in V(G_A^{nt})\setminus \{v_i\}$ where $V(G_A^{nt})$ is the node set of $G_A^{nt}$. Nodes with top-k similarities are connected to $v_i$, so the edges $e_{ij}$ are generated through:
\begin{equation}
	e_{ij} = 1 \quad if \ \text{TiSim}(v_i, v_j) \in \text{top-k}\{TiSim(v_i, v_n)\}.
\end{equation}
Here TiSim$(v_i,v_j)$ is the time-series similarity function between nodes $v_i$ and $v_j$ like the inner product, and empirically we select $5\%$ of the total number of nodes as k. 

To characterize each node with fine-grained features, we project the input into a $d_A$-dimensional space with learnable parameters $W_A$. That is to say, the node feature dimension on the attribute graph is $d_A$. Given a node $v_i$ that corresponds to an original attribute $attr_i$, the impact of attribute-attribute pair varies among neighbor nodes $\mathcal{N}(v_i)$. For example, "delivery\_time" is more correlated with "delivery\_distance"
other than "pay\_amount." In order to distinguish the importance of different neighbors, we utilize graph attention networks\cite{velivckovic2017graph} which have shown remarkable capacity on graph data. Generally, the graph attention network computes the latent representations $H_i$ of $v_i$ by:
\begin{equation}
	H_i=GAT(G, Z_i)=\sigma(Z_i+\sum_{v_j\in\mathcal{N}(v_i)} \alpha_{ij} Z_j) 
\end{equation}
where $G$ is the graph, $Z_i$ is the input attribute $attr_i$ and $\sigma$ is the sigmoid function. $\alpha_{ij}$ is the attribute attention score measuring how important $v_j$ is to $v_i$. The first term denotes the self-message and the second term means information aggregated from neighbors. The attribute attention score $\alpha_{ij}$ is calculated by:
\begin{equation}
\begin{split}
	e_{ij}'&=LeakyReLU(W_e[Z_i||Z_j]) \\
	\alpha_{ij}&=\frac{e'_{ij}}{\sum_{v_n\in \mathcal{N}(v_i)} exp(e_{in}')}.
\end{split}
\end{equation}
Here $[\cdot||\cdot]$ denotes the concatenation operation. $W_e\in\mathbb{R}^{2d_A}$ are learnable parameters and $LeakyReLU$ is the activation function. As there are stable component and volatility component, we can obtain their representations by $H_i^V=GAT(G^{nt}_A, X_i^VW_A)$ and $H_i^S=GAT(G^{nt}_A, X_i^SW_A)$, respectively. Note that the parameters of the two GAT are not shared.

\subsection{Entity-Temporal Graph Attention Network}

Apart from the attribute relations, another perspective that needs to be considered is the relations between entities under the dynamic settings, because information that individual entities obtain can be enhanced through passing messages among correlated entities. For example, one retailer can judge whether there is an anomaly in the transaction by comparing his own operating conditions with similar retailers in the scenario of Ele.me. Motivated by this idea, we capture such relations based on static and dynamic aspects. Firstly, entities with similar static properties are supposed to be more correlated. Unlike desserts suppliers, retailers selling spicy dishes will share similar operating conditions with another spicy food supplier. Therefore, the pre-extracted static attributes are utilized to construct an entity graph $G_E$ the same way as the attribute graph $G_A$. Besides, it is natural that entities with higher temporal pattern similarities are more correlated. For instance, the transaction of a newly started retailer varies like other incipient retailers and is distinguished from retailers that have been opened for a long time. Thus, we calculate the temporal graph $G_T$ based on the time-series of attributes across entities. The entity-temporal correlations among entities are captured through the entity-temporal graph, of which the adjacency matrix can be formulated as:
\begin{equation}
	\tilde{A}=\left[
		\begin{array}{cc}
			A_E & A_T \\ 
			A_T & A_E
		\end{array}
	\right]
\end{equation}
where $A_E$ and $A_T$ is the adjacency matrix of entity graph and temporal graph, respectively. Graph neural network on entity-temporal graph at two adjacent timesteps can simultaneously extract information from two aspects, which is formulated as:
\begin{equation}
	\tilde{H}^t=\tilde{A}\left[\begin{array}{cc}H^{t-1}\\ H^{t}\end{array}\right]\tilde{W}=\left[
		\begin{array}{cc}
			A_E & A_T \\ 
			A_T & A_E
		\end{array}
	\right]\left[\begin{array}{cc}H^{t-1}\\ H^{t}\end{array}\right]\tilde{W}.
\end{equation}
The above equation can be rewritten as:
\begin{equation}
    \tilde{H}^t = A_E H^{t-1}\tilde{W}+A_E H^{t}\tilde{W}+A_T H^{t-1}\tilde{W}+A_T H^{t}\tilde{W},
\end{equation}
which indicates that it is the combination of 4 independent GNNs. $H^0$ is the output of attribute graph attention network and it can be obtained from stable or volatility component. Given the neighbor difference of central entities, we replace GNN with graph attention networks and the representation at current step $t$ is transformed into:
\begin{equation}
	\tilde{H}^t=\sum_{k=t-1}^t GAT(G^*, H^k), \quad G^*\in\{G_E,G_T\}.
\end{equation}

\subsection{Anomaly Detection}
\label{mutual_compute}
The representations from the stable component and the volatility component are concatenated for a MLP layer to generate prediction of future KPI $\hat{Y}$. We use MSE loss as the objective to optimize GraphAD as $L_{MSE}=\frac{1}{n}\sum_i^n ||\hat{Y}_i-Y_i||^2$. Given that $X=X^V+X^S$, Eq. \ref{eq:stable} and \ref{eq:vola} can be unified and the overall objective can be formulated as:
\begin{equation}
    \label{eq:loss}
   \min L=L_{MSE} + \lambda I(X,X^V).
\end{equation}
$\lambda$ is the hyper-parameter for the mutual information-based regularization term. However, it is of great difficulty to compute mutual information in high-dimension, so we adopt CLUB \cite{cheng2020club} to minimize the upper bound of $I(X,X^V)$ as an approximation. Note that if the gap between prediction and ground-truth is larger than that on the training set, we will judge the prediction as anomalies.
\section{experiment}

\subsection{Dataset}
\subsubsection{Background}
We construct a dataset from transaction data in Ele.me, a representative scene for entity-wise multivariate time-series anomaly detection. Raw data consist of transaction records between retailers and consumers from Jun. 30th, 2020 to Jun. 30th, 2021, containing desensitized information about consumers, retailers and couriers. There are 70 attributes in one record, including order details like pay amount, delivery time and distance. 
\subsubsection{Preprocessing}
To avoid the time alignment problem caused by missing records on different days for various retailers, we choose retailers that sell goods every day during the selected duration. Besides, anomaly detection for retailers with low turnover is worthless from the industry perspective, so we filter out retailers whose average daily transaction records are less than 20.

The data are organized as $X\in\mathbb{R}^{N\times T\times D}$ where $N$ is the number of retailers, $T$ is the number of days and $D$ is the attribute dimension. We aggregate numeric attributes of daily records in average or summation form, while category attributes are counted by values. We also take into account the impact of couriers; for example, couriers delivering food at a lower speed are more likely to raise transaction issues. Hence, we design couriers-related attributes like "number of order delivery on time." Finally, we obtain 114 attributes for each retailer each day. The corresponding labels are generated based on $X$ and business requirements.

Given the static similarity between retailers, we also extract static retailer attributes such as "open time", "product type", and "location". Furthermore, our goal is to use 30 days of data to decide the operating condition on the 31st day.

\subsubsection{Data Distribution}
We detail the dataset to provide intuition on the scene of Ele.me. From the perspective of delivery, the number of orders sent to consumers ahead of time follows the Gaussian distribution, while the overtime orders follow a uniform distribution. The total number of abnormal days varies across retailers, with the majority being three. The number of retailers with a total number of abnormal days less than 7 takes up more than 80$\%$ of the data, revealing the long-tail distribution of the dataset.

\subsection{Experiment Setup}

The data are firstly normalized and we apply a sliding window of 31 days to generate samples. We then split the data into train/validation/ test set with a ratio of 6:2:2.
We utilize normal data to train GraphAD, and the largest difference between prediction and ground-truth will be the threshold for validation and testing. Note that the threshold also varies across entities. The model with the best performance on the validation set is selected to conduct the final test. Precision, Recall, F1-score and AUC(ROC) are the metrics measuring model effectiveness. We use a random search strategy to find the optimal hyper-parameters, and Adam \cite{kingma2014adam} is adopted for model optimization. The optimal hyperparameter $\lambda$ in Eq. \eqref{eq:loss} is 12.41 and the learning rate is 5$\times 10^{-6}$. For comparison, we choose AE, VAE \cite{an2015variational}, IsolationForeset \cite{liu2008isolation}, DeepSVDD \cite{ruff2018deep} and COPOD \cite{li2020copod} as the baseline methods. Note that although GDN \cite{deng2021graph} and MTAD-GAT \cite{zhao2020multivariate} are both graph-based methods, they can only be trained on a single entity. It will be extremely time-consuming if we train models on each entity one by one. Besides, the variance of entity features will cause the model to perform well only on the current entity and the performance would be degraded on other entities. Based on these two reasons, we do not choose them as baseline methods.

\subsection{Evaluation Result}
The performance of GraphAD and other baseline methods are organized in Tab.\ref{tab:basic}. We can observe that GraphAD achieves the best performance on Precision, F1-score and AUC, and it also obtains a competitive result on Recall. A relative improvement of 184.4\% is achieved in the Precision, and 62.6\% in F1-Score, which demonstrates the superiority of our designed model. Note that in industrial scenarios, reporting true anomalies is more important than reporting many false anomalies for it may mislead users. Therefore, although Recall of GraphAD is slightly at a disadvantage, its significant improvement in precision can bring significant economic benefit, which is much more valuable to Ele.me.

\begin{table}[!h]
\centering
\begin{tabular}{c|cccc}
\toprule
\textit{\textbf{Model}} & \textit{\textbf{Precision}}   & \textit{\textbf{Recall}}     & \textit{\textbf{F1-Score}}   & \textit{\textbf{AUC}}        \\ \hline
AE                      & 19.02  & 35.00  & 24.65  & 63.60   \\
VAE                     & 20.11  & 37.00  & 26.06  & 64.46  \\ 
IsolationForest         & 26.67  & \textbf{44.00} & 33.21  & 68.83  \\ 
DeepSVDD                & 13.71  & 27.00  & 18.18  & 59.05  \\ 
COPOD                   & 22.70  & 42.00  & 29.47  & 67.26  \\ \midrule
\textbf{GraphAD}       & \textbf{75.86} & 41.90  & \textbf{53.99} & \textbf{70.58} \\ \bottomrule
\end{tabular}
\caption{Performance comparison on Ele.me dataset.}
\label{tab:basic}
\end{table}

\subsection{Ablation Study}
To verify the effectiveness of each designed module, we also conduct an ablation study of GraphAD. In more detail, we remove the K-Decom module and directly use the original KPI as the input for the latter modules, which is denoted as "w/o K-Decom". Similarly, we also remove the attribute GAT, entity GAT and temporal GAT, and the corresponding models are named after "w/o A-GAT", "w/o entityGAT", "w/o temporalGAT", respectively. We organize their performance on Tab. \ref{tab:ablation}. Performance of the model without distinguishing the entity difference degrades most, revealing the importance of modeling from the perspective of entities. All metrics of other variants are also decreased, which indicates the reasonable design of each module.

\begin{table}[!h]
\centering
\begin{tabular}{c|cccc}
\toprule
\textit{\textbf{Model}} & \textit{\textbf{Precision}}   & \textit{\textbf{Recall}}     & \textit{\textbf{F1-Score}}   & \textit{\textbf{AUC}}        \\ \hline
\textbf{GraphAD}& \textbf{75.86} & \textbf{41.90} & \textbf{53.99} & \textbf{70.58} \\\hline
w/o K-Decom  & 70.91          & 37.14          & 48.75          & 68.15            \\
w/o A-GAT     & 61.40          & 33.33          & 43.21          & 66.09          \\
w/o entityGAT   & 53.85          & 26.67          & 35.67          & 62.70          \\
w/o temporalGAT  & 67.92 & 34.29 & 45.57 & 66.70 \\\bottomrule
\end{tabular}
\caption{Ablation study of GraphAD}
\label{tab:ablation}
\end{table}

\section{conclusion}

We pose the entity-wise multivariate time-series anomaly detection problem in the scenario of the O2O platform. We incorporate the impact of different components of KPI and model the attributes, entities and temporal dimensions as graphs. The proposed GraphAD is proven to be effective on our constructed dataset from real transaction data in Ele.me.

\clearpage
\bibliographystyle{ACM-Reference-Format}
\bibliography{ref}

\end{document}